*CCC White Paper*
**October 2026**

Computing Community Consortium
Visioning

# Securing the Robotics Supply Chain

*Technical Needs and Investment Opportunities for the Domestic Robotics Research Ecosystem*

**Authors**

Cindy L. Bethel (University of Alabama), Henrik I. Christensen (University of California San Diego), Odest Chadwicke Jenkins (University of Michigan), George Pappas (University of Pennsylvania), Holly Yanco (University of Massachusetts Amherst), and Haley Griffin (Computing Research Association)

## Executive Summary

On July 28, 2026, the United States (U.S.) Federal Communications Commission (FCC) added two categories of foreign-produced devices to its Covered List — advanced robotic devices, including mobile humanoids and quadrupeds, and connected power inverters — closing off new United States equipment authorization for these products (Federal Communications Commission, DOC-423682A1, 2026). The action responds to a substantive security concern: robotics supply chains pass through manufacturers the U.S. cannot fully vet, networked robots present a plausible vector for surveillance or remote control, and critical infrastructure increasingly depends on hardware assembled from components of uncertain provenance. Modern mobile robots are sophisticated networked systems, equipped with cameras, sensors, onboard computers, and wireless communications, that can observe and record the factories, warehouses, infrastructure, homes, and public spaces in which they operate; ensuring that this operational and environmental data cannot be transmitted to unauthorized foreign entities is a legitimate national priority.

The premise underlying the ruling merits serious attention. Less visible outside the laboratories and companies building these systems, however, is how extensively the current U.S. robotics research base already depends on the very supply chains the ruling is designed to restrict. This situation is not because researchers prefer foreign built robots, but because domestically manufactured alternatives remain scarce and, where they exist, priced well beyond a typical research budget. This dependence extends well past finished robots and down to its core components: the motors, sensors, and wiring inside robots assembled domestically are still, in large part, manufactured abroad.

The stakes extend well beyond the research ecosystem alone. Robotics is arguably experiencing an "iPhone moment": the convergence of increasingly capable mechanisms, powerful and affordable computing, and modern artificial intelligence is transforming what robots can do and where they can operate, with implications across service robotics, logistics, manufacturing, agriculture, healthcare, infrastructure, and defense. The same affordability pressures facing robotics researchers also reach public safety agencies, where drones used for wildfire detection and other emergency response functions can become unaffordable well before a domestic alternative exists to replace them (Stockman, 2026). Restricting access to robotic technology before competitive domestic alternatives exist risks raising automation costs precisely when U.S. companies need automation to address labor constraints, improve productivity, and compete globally, and risks slowing the broader reshoring of American manufacturing and modernization of logistics that robotics increasingly makes possible.

This paper sets out what a sustained federal research and investment agenda would need to include to close that gap — not within a single budget cycle, but over the decade such an undertaking will realistically require. It identifies where the gap is most pronounced today: in the methods needed to verify and certify components of uncertain origin, in the foundational hardware research that has long been underfunded relative to software, in the physical infrastructure researchers need to test systems at scale, and in the workforce required to build and sustain all of it. Robotics should not be treated merely as a downstream beneficiary of a broader artificial intelligence investment strategy. It constitutes its own supply chain, with its own points of failure, and merits its own sustained federal commitment.

## Background: The Ruling

The Covered List addition blocks new equipment authorization for the import, marketing, or sale of two device categories in the U.S.: advanced robotic devices (including mobile robots, humanoids, and quadrupeds) and connected power inverters, effective July 28, 2026. Previously authorized models and existing consumer purchases are unaffected, federal government use is exempt, and a conditional-approval pathway exists through Department of Defense and Department of Homeland Security review.

The Commission's determination cites supply chain vulnerability that could "disrupt U.S. economic and national security," cybersecurity risk to critical infrastructure, and the possibility that networked robots could be used to conduct surveillance, strengthen foreign intelligence capabilities, or be remotely commandeered. These are not hypothetical risks. Consumer robots have reportedly transmitted video data to third countries without users' knowledge (Sabin, 2025); autonomous vehicles have allegedly been controllable from abroad (Bureau of Industry and Security, 2025); and components have been withheld from U.S. buyers amid unrelated geopolitical disputes (Bry, 2024).

Enforcement raises further questions, including how the rule will treat firms that assemble domestically while remaining foreign-owned, or that market themselves as domestic while depending on foreign supply chains several tiers removed. The restriction's reach is also broader than dependence on international suppliers alone. It may affect a wide range of advanced foreign-made mobile robotic systems generally, while many traditional-developed industrial manipulators continue to be treated differently under existing rules.

It is crucial that the outcome of this ruling is a robotics industry the U.S. can genuinely trust rather than one that simply excludes a defined list of devices. This ruling represents a starting point, not an endpoint. Strengthening the broader robotics research and manufacturing ecosystem is a task only sustained federal investment is positioned to carry out; individual companies and universities each hold only a fragment of the picture, and no single actor commands the resources or mandate to rebuild an entire supply chain independently. This paper lays out a technical research agenda and the categories of federal investment that would allow the U.S to meet the security objectives behind this ruling without leaving its own researchers, and its own industry, without the means to compete.

## The Current State of the Domestic Robotics Research Ecosystem

Unitree, China's leading humanoid and legged robotics company, has become the dominant supplier of humanoid research robots. It shipped more than 5,500 humanoids in 2025, over a quarter of the global market, and 74 percent of its humanoid sales go to universities, research institutions, and individual developers; only 9 percent go to industrial applications (Campbell, 2026). The driver is cost. Over six years Unitree has cut the price of its quadrupeds from $45,000 to under $2,000, and its flagship G1 humanoid now sells for about $13,500 (Campbell, 2026). By comparison, Boston Dynamics' Spot quadruped and became commercially available at a base price of $74,500 (Bray, 2020). A research lab can therefore field a fleet of Unitree robots for the price of a single comparable domestic platform.

The comparison illustrates the cost gap, but not a clean domestic-versus-foreign contrast. Boston Dynamics, though headquartered and substantially operating in the U.S, has been a wholly owned subsidiary of South Korea's Hyundai Motor Group since mid-2026, the same kind of domestically-based-but-foreign-owned structure whose treatment under the ruling remains unresolved. Few U.S.-manufactured research robot platforms exist at all, and those that do remain priced beyond the reach of most academic laboratories and could also be outside the viable price point for many small and mid-size businesses. At a time when university research budgets are already under significant pressure, sharply reducing access to affordable research platforms risks making it harder for American researchers and students to innovate at the pace of their international counterparts. Furthermore, restricting access to needed technology ahead of an affordable domestic alternative does not eliminate the need for that technology; it simply shifts the cost onto the researchers, agencies, and communities that depend on it.

The U.S. research community dependence does not end with finished robots. Even robots assembled domestically rely on foreign-sourced components: sensors, wiring, and, in particular, motors. The actuators that literally move a humanoid robot are, at present, not produced (or "wound") in the U.S.. Components designed domestically typically must still be manufactured overseas. Batteries, semiconductors, cameras, and embedded compute platforms show the same pattern, frequently originating outside the U.S. regardless of where a given robot is ultimately assembled. This dynamic also suggests a possible advantage for American developers: international manufacturers are highly effective at producing components cheaply, but those components are generally not co-designed with the software and control systems that operate them. **Integrated hardware-software co-design**, rather than unit cost, is an area in which U.S. developers could differentiate themselves.

The reliance on international goods related to robotics has already proven fragile in practice. Components have been withheld from U.S. buyers amid unrelated geopolitical disputes (Kesteloo, 2024). The multi-year effort to scale domestic drone production and procurement, dating to 2020 and still incomplete, illustrates how slowly "reshoring" a hardware supply chain actually proceeds. Chip supply chains for otherwise unrelated hardware have reportedly been traced to a small number of internationally-linked suppliers several tiers removed (Saunders, 2026).

The U.S. is not starting from parity. China remains the leader in robot density, ahead of the U.S. — 322 robots per 10,000 workers, compared with 274 in the U.S. The volume of national robot sales in China runs at roughly four times the level in the U.S. (CCC, Robotics Mid-Cycle White Paper, 2023, Sec. 2.1, p.3). As the Computing Community Consortium's (CCC) 2023 mid-cycle assessment observes, "Already today, the US is falling behind other nations in terms of basic research" (CCC, Robotics Mid-Cycle White Paper, 2023, p.2). More broadly, several Asian nations are pairing this market lead with large domestic markets that let companies deploy systems at scale, accumulate operational experience, reduce costs, and iterate rapidly, reinforced by coordinated long-term national robotics strategies and substantial public and private investment.

The greater strategic risk to the U.S. is therefore not simply current dependence on foreign-made robots, but the possibility that the next generation of robotics technology, manufacturing capability, supply chains, and engineering expertise becomes concentrated elsewhere. Absent a deliberate transition plan, withdrawing researchers' access to the low-cost foreign platforms they currently rely upon risks widening, rather than narrowing, this gap.

## Technical Research Priorities for a Secure, Domestically Capable Ecosystem

**Designing for a mixed-trust supply chain.** For the foreseeable future, U.S. robotics systems will be built, at least in part, from components whose full provenance cannot be verified. A pure computing stack is likely unattainable. Treating this as a temporary problem to be resolved through import restriction alone risks isolating U.S. researchers from a field that will continue to advance regardless.

A more durable approach is a research agenda aimed at building systems that can be trusted even where the underlying supply base remains partly foreign for a decade or more. Security could be a luxury that US researchers can no longer afford. The country will have to operate with components it cannot fully trust, which makes verification, rather than exclusion, the central task. That task extends to analysis as well as verification: the ability to reverse-engineer a component and determine whether it carries embedded malicious functionality, and where necessary disable or remove it, is not yet a well-established discipline within cybersecurity as generally practiced, and building it will require both new technical methods and a trained workforce.

A genuine tension is worth acknowledging here rather than setting aside: restricting access to foreign-made platforms also limits researchers' ability to study how those platforms operate, which is itself a precondition for developing the analysis capability this problem demands. The U.S. already manages a version of this problem in other industries, like small electronics (mobile devices, laptops etc.) and that same discipline needs to be extended to robotics.

**Designing robots to tolerate supplier substitution.** Building a domestic supply base will necessarily mean shifting suppliers and varying specifications for years, as manufacturing capacity is developed. This argues for research into formal interfaces that specify precisely what a robot's control software depends on from a given component, automated methods for re-identifying system behavior once a component has changed, techniques for transferring learned control policies within bounded performance guarantees, and architectures designed to minimize the cost of safety revalidation each time a substitution occurs.

**Investing in the hardware fundamentals: *Characterizing what component quality actually requires*.** Specifications in robot manufacturing can conflate supplier-specific design choices with genuine functional requirements, treating a component as essential when it is merely an artifact of a particular supplier's design. Wherever that turns out to be the case, it represents a manufacturing constraint that can be relaxed, potentially widening the pool of qualified domestic suppliers. Little empirical work has established how much mechanical backlash, torque ripple, or sensor degradation a robot's control software can actually absorb before task performance deteriorates, a threshold that has never been systematically mapped despite being measurable today on existing hardware.

This argues for sustained investment in the hardware fundamentals that characterization work depends on: component-level research into actuators, grippers, sensors, new materials, and fully sensorized end-effectors, which has been persistently underfunded relative to software and systems research (CCC, Robotics Mid-Cycle White Paper, 2023, Sec. 3.2, pp.7, 13), and which CCC's 2020 roadmap already flagged as a critical, underinvested area spanning gripper technology, multi-modal sensor integration, and advanced manufacturing methods (CCC, Robotics Roadmap 2020, Sec. 2.2/3.2, pp.18, 37- 38).

**Building the tools to verify trust, not just assume it.** There is currently no established, research-backed method to validate and certify the safety and security of a robotics component or system before it is put into service, regardless of where it was made. Current methods also make it difficult to safely certify systems that rely on deep learning in safety-critical control loops, a limitation directly relevant to certifying any system built from a mix of trusted and untrusted parts (CCC, Robotics Mid-Cycle White Paper, 2023, Sec. 3.2, p.7).

Addressing this gap means replacing proxies based on a component's country of origin with testable trust properties: machine-readable hardware and software bills of materials that resolve down to individual logic-bearing subcomponents, model provenance tracking, signed update channels, and hardware attestation, each verifiable regardless of where a device was manufactured. Such an approach must also account for a transitive problem, since foreign suppliers frequently depend on parts made outside their own countries which are several tiers down their own supply chains, meaning diversification at only the first tier of suppliers can be misleading. In discussion, certification and validation were identified as the key gap precisely because there is currently no established way to do this, and it was proposed that verification be reframed around outcomes rather than origin: rather than determining whether a component should be used at all, the more productive question is what a given component needs to be verified to do, testing on the assumption that it could be adversarial until proven otherwise.

**Funding shared, open-access test infrastructure.** A fully instrumented, multi-robot research setup can easily cost $1 million, a barrier that puts serious experimental robotics work out of reach for many laboratories (CCC, Robotics Mid-Cycle White Paper, 2023, Sec. 4, p.13; CCC, Robotics Roadmap 2020, Sec. 7, p.79). Open-access national test facilities, spanning manufacturing, warehousing, home care, surgery, multi-robot systems, drones, and agriculture, would lower that barrier considerably, and this kind of shared infrastructure would become more urgent, not less, if researchers lose the option of falling back on inexpensive imported platforms as a stopgap.

## Federal Investment Opportunities Across the Robotics Supply Chain

**A conditioned pathway for research access.** Academic research use is not currently an enumerated condition for equipment authorization under 47 CFR § 2.1204, and as previously authorized platforms lose their path to authorization, U.S. research institutions are left with no clear route to obtain newer models for research purposes. A registered, conditioned import pathway limited to qualified institutions (covering modest quantities, network isolation, no resale, defined inventory and disposition obligations, and vulnerability reporting) would preserve research access without reopening the door the ruling is meant to close.

**Targeting the components that actually constrain supply, not final assembly.** Domestic incentives directed only at final robot assembly will not reach the true chokepoints in the supply chain: motors and actuators, drives, compute, sensors, precision reduction gearing, wiring harnesses, and batteries. Meaningful domestic-content thresholds will remain unreachable without direct investment at this layer, since incentives aimed only at assembly will not reach it. The objective of this investment should be explicit: establishing domestic production capacity for these components at prices that are globally competitive, not merely domestically available.

**Supporting an open, manufacturable reference design.** No U.S. robotics company currently has a maintained, open reference design (covering mechanical design, bill of materials, firmware, control software, and a validation suite) that it can build on rather than re-solving from first principles. That repeated fixed cost is itself part of why there are relatively few domestic robotics manufacturers rather than many. This extends naturally to promoting open-source and open-hardware development more broadly, as something the federal government can support beyond the purely technical reference-stack.

**Closing the gap between federally funded research and commercial products.** Federally funded robotics prototypes frequently fail to reach the domestic market, even when the underlying research was domestically conducted. Army Research Laboratory-funded quadruped research is a case in point: it was ultimately commercialized abroad, and the resulting domestic product now sells at roughly sixty times the price and a fraction of the volume of the version developed elsewhere (Martina, 2026). Expanding NSF's Translation to Practice (TTP), I-Corps, and SBIR/STTR pathways specifically for robotics hardware would help close this gap, building on CCC's existing recommendation along the same lines (CCC, Robotics Mid-Cycle White Paper, 2023, Sec. 5, p.14).

This is a two-directional gap, not only a one-way "lab to market" problem: it is also a "market to lab" problem, since some components are manufactured abroad and cannot yet be produced domestically at all, even in small research quantities. This is especially true for basic motor components. There is potential for newer translation-focused federal initiatives, or an NSF program focused specifically on lab-to-market translation, to have a natural robotics analog worth pursuing.

Closing this gap should not come at the expense of basic research at the exploration side of the Technology Readiness Level (TRL) spectrum for forming new innovations (NASA, 2023). University robotics research characteristically sits between TRL 1-4 in laboratory environments, and requires a different but complementary aptitude for the translation work to real world application that spans TRL 5 through 7 for products and deployments that result from TRL 8-9 work. Treating that gap as a shift of the universities and research to higher TRL work rather than an expansion across the TRL spectrum risks crowding out the wellspring of new ideas responsible for transformative innovations.

Decades of federally funded basic research, much of it without any obvious near-term commercial application, preceded the Internet, artificial intelligence, cryptography, and many other capabilities regularly used today (National Academies of Sciences, Engineering, and Medicine, 2020; Nilsson, 2010). A domestic robotics ecosystem built only to accelerate what is already close to market would miss the open-ended, lower-TRL research most likely to transform the field over the next several decades. The two priorities are complementary, not competing: closing the lab-to-market gap for higher-TRL work, while preserving room for exploratory basic research whose payoff has not yet declared itself, are both necessary parts of the same investment agenda.

**Coordinating with, rather than duplicating, efforts already underway.** The Special Competitive Studies Project's newly launched National Security Commission on Robotics for Advanced Manufacturing, announced in March 2026, is organized around four pillars directly relevant to this paper: a unified public-private investment strategy, a domestic talent pipeline, measurable integration benchmarks, and securing U.S. leadership across hardware, software, and supply chains, with a final report due in March 2027 (Special Competitive Studies Project, 2026).

**A coordinated federal approach, not a single-agency mandate.** No single agency, and no single division within an agency, is positioned to address every dimension of this problem on its own. What is needed is a genuinely coordinated federal effort rather than a proliferation of separate ones: a multi-agency program spanning fundamental research, component technology, systems integration, and security considerations, aligned with the SCSP Commission's forthcoming report, and paired with published Conditional Approval criteria and timelines so that domestic firms can plan accordingly. This coordination extends beyond funding allocation alone. Regulatory choices, economic incentives, and procurement policy all shape outcomes in this space, at times more directly than direct grant funding does; a whole-of-government approach that draws on all of these levers, rather than funding alone, is likely to prove more effective than a purely multi-agency funding model.

Any structure that incorporates corporate funding partners should be designed deliberately, with clear terms for intellectual property sharing and publication independence built in from the outset, so that the resulting institute remains sustainable and its research agenda stays aligned with the broader national interest even as it draws on industry's resources and expertise. Timing matters as much as structure: restrictions calibrated to phase in as domestic alternatives become available and affordable, rather than in advance of them, would give researchers and manufacturers alike a predictable transition rather than an abrupt one.

## Workforce and Education as Supply Chain Infrastructure

A domestic robotics supply chain is, in the end, as much a workforce need as a manufacturing one, if not more. Highly skilled and trained citizens is the primary investment made by federal funding. If the U.S. moves toward relying primarily on trusted, domestically built platforms, it does not possess the trained workforce to support that shift today. In fact, the individuals who would eventually build that capacity are, in large part, currently in grade school. Closing this gap means building capacity across the entire pipeline, from K-12 through the doctorate, and it is not primarily a question of producing more college graduates; technical schools, community colleges, and undergraduate programs are as essential to this workforce as PhD-granting institutions. This is a multi-decade undertaking, which argues for treating education investment as central to this agenda rather than a secondary concern.

The new NSF UIDP Industry-Integrated Ph.D. Scholars Program offers approximately $47 million to support more than 250 students over five years on a four-year, applied-practice doctoral track that includes a required one-year industry placement (National Science Foundation, 2026). This program is a working precedent for applied, industry-facing doctoral training. In its current form, however, the NSF UIDP remains modest relative to the scale of the need and may be too limited to generalize. A genuine design tension exists here: a four-year timeline is often too short for authentically hands-on training, and heavy reliance on corporate placements can create the same intellectual-property and sustainability concerns noted above in the discussion of corporate research partnerships.

Partnering with national laboratories, rather than exclusively with companies, is one way to preserve the applied-training benefit while reducing that risk. A complementary model worth evaluating is a professional doctoral track comparable to Europe's “industrial PhD” that would give students interested in applied, industry-relevant work a credentialed path higher up the technology readiness spectrum, without displacing the traditional research doctorate for students focused on basic research (Department of Informatics & Telecommunications, University of Ioannina, n.d.). NSF is reportedly already piloting a version of this model and its results should inform how the U.S. broadens applied doctoral training going forward.

The scale of the competitive gap is also relevant here. Whatever the U.S. commits to training and funding in this space is likely to be substantially outnumbered by comparable investment abroad, with direct downstream effects on manufacturing and research competitiveness. The growth in international professional society membership and international enrollment in U.S. graduate programs must be balanced against building a domestic pipeline. However, domestic student interest in graduate study in this field is limited in part because industry compensation is significantly higher.

Compensation is one part of this, but the constraint may also be a weak connection between where interest already exists and where it needs to go. K-12 robotics competitions, such as FIRST Robotics, generate substantial early interest in the field nationally; at the University of Michigan, the demand created by this kind of K-12 pipeline was itself a significant factor in the university's decision to establish a dedicated Robotics major and department, which drew more than 100 declared majors in its first year alone (Jenkins et al., 2023). Better connecting K-12 competition programs to higher education, rather than assuming interest itself is the bottleneck, could meaningfully expand the domestic talent pipeline. Building that connective tissue, from K-12 competitions through vocational and graduate programs alike, is likely to do more than incentives aimed at graduate study alone.

The challenge is that this is not solely a funding question for the federal government. Universities themselves have room to rethink parts of their own training model rather than treating this entirely as something government investment alone should resolve. Universities often view robots as abstractions at a conceptual level, but could try to realize a better balance of theory and practice (as well as teaching and research). One structural issue worth addressing directly: U.S. academic robotics research has tended to treat robot platforms as commodities, with research conducted "on top of" them, which has separated researchers doing AI and software work from those doing manufacturing and hardware work. Rebuilding a genuine domestic supply chain requires addressing that separation directly. Building on top of unreliable or untrusted systems, manufacturing tolerance, and higher-level AI methods are, in a real sense, the same category of research problem, and none of it succeeds if the underlying disciplinary silos remain in place.

Practically, this argues for training that returns to first principles: coursework and hands-on experience that let students build the fundamental components of a robot (actuators and motors, for instance) rather than only programming atop platforms supplied by companies. Universities already operate microfabrication facilities so that engineering students understand, through direct experience, what it takes to design and produce a chip. A comparable shared facility for robotics, paired with training built on top of it, would extend that same first-principles model to robotics hardware. That same model is largely absent below the university level, however. Very few technical schools or community colleges currently offer robotics-specific training of any real breadth, even though the technicians such programs would produce are as necessary to a functioning domestic supply chain as researchers with doctorates. Such training, at every level, should build security into component design from the outset, rather than treating it as something appended after a system is otherwise complete. This security by design constitutes better engineering practice, and addresses a security concern that extends well beyond national security to the basic question of whether a robot can be remotely compromised at all.

Taken together, these threads point to an undeniable opportunity, not simply a problem to be managed. Closing the robotics workforce gap will not come from PhD-level investment alone, or from federal funding alone. It requires sustained commitment across the entire pipeline, from K-12 through the doctorate, spanning technical schools and community colleges as much as research universities, and it requires universities to close the disciplinary gap that has long separated those who build software from those who build hardware. Training built this way would instill hardware-software co-development from the beginning of a system's design, combined with security built in from that same starting point rather than patched on later. An ecosystem organized to support that kind of integrated development, at scale and sustained over the decades such a transformation actually requires, would represent a durable U.S. advantage.

## Conclusion

The security concerns underlying the Commission's ruling identify a genuine vulnerability, and that vulnerability being raised at the national level is an occasion for a larger opportunity: to build, deliberately and at scale, the domestic robotics research and manufacturing base the U.S. has so far lacked. Seizing that opportunity means closing a real gap, one in which dependence on foreign platforms and components runs deep, and in which a domestic alternative does not yet exist at meaningful scale, through sustained, coordinated investment rather than a single policy action.

The research priorities and investment opportunities set out above, spanning component-level hardware research, methods for verifying trust within a mixed supply chain, and shared physical infrastructure for testing at scale, constitute the technical foundation of that investment. None of it can be realized, however, without an equally sustained commitment to workforce and education: the researchers, engineers, and technicians capable of designing, certifying, and manufacturing trustworthy robotic systems will not exist on the timeline this problem demands unless that training begins now, at every level from foundational hardware instruction through doctoral education.

Pursued together, and with the ambition this moment allows, these priorities offer the U.S. a genuine opportunity to move ahead of its competitors rather than simply close the distance to them — building a robotics research and manufacturing base it can lead with, not merely depend upon.

## Suggested Citation

Bethel, C. L., Christensen, H. I., Jenkins, O. C., Pappas, G., Yanco, H., & Griffin, H. (2026). *Securing the robotics supply chain: Technical needs and investment opportunities for the domestic robotics research ecosystem.* Washington, D.C.: Computing Research Association (CRA). http://cra.org/ccc/wp-content/uploads/sites/2/2026/10/Securing-the-Robotics-Supply-Chain.pdf

## Acknowledgements

### Reviewers

CCC and the authors acknowledge the reviewers whose thoughtful comments improved the white paper:

- **Nenad Medvidović,** University of Southern California
- **Weisong Shi,** University of Delaware

*This material is based upon work supported by the U.S. National Science Foundation (NSF) under Award Nos. 2300842 and 2619366. These awards support the Computing Community Consortium (CCC), a programmatic committee of the Computing Research Association (CRA).*

*Any opinions, findings, and conclusions or recommendations expressed in this material are those of the author(s) and do not necessarily reflect the views of the National Science Foundation.*

### About the Computing Community Consortium (CCC)

A programmatic committee of the Computing Research Association (CRA), CCC enables the pursuit of innovative, high-impact computing research that aligns with pressing national and global challenges. Of, by, and for the computing research community, CCC is a responsive, respected, and visionary organization that brings together thought leaders from industry, academia, and government to articulate and advance compelling research visions and communicate them to stakeholders, policymakers, the public, and the broad computing research community.